\documentclass[sort&compress,preprint]{elsarticle}

\usepackage{amssymb, amsmath, amsthm, amssymb, amsfonts}
\usepackage{algorithm,algorithmic}
\usepackage{multirow,subfigure}
\usepackage{cleveref}
\usepackage{url}

\newtheorem{defn}{Definition}

\journal{Information Sciences}

\begin{document}

\begin{frontmatter}




\title{Fair Oversampling Technique using Heterogeneous Clusters}


\author[inst1]{Ryosuke Sonoda}
\ead{sonoda.ryosuke@jp.fujitsu.com}

\affiliation[inst1]{organization={Fujitsu Ltd.},
            addressline={4-1-1 Kamikodanaka, Nakahara-ku}, 
            city={Kawasaki-shi},
            postcode={211-8588}, 
            state={Kanagawa},
            country={Japan}}


\begin{abstract}
Class imbalance and group (e.g., race, gender, and age) imbalance are acknowledged as two reasons in data that hinder the trade-off between fairness and utility of machine learning classifiers.
Existing techniques have jointly addressed issues regarding class imbalance and group imbalance by proposing fair oversampling techniques. 
Unlike the common oversampling techniques, which only address class imbalance, fair oversampling techniques significantly improve the abovementioned trade-off, as they can also address group imbalance.
However, if the size of the original clusters is too small, these techniques may cause classifier overfitting.
To address this problem, we herein develop a fair oversampling technique using data from heterogeneous clusters.
The proposed technique generates synthetic data that have class-mix features or group-mix features to make classifiers robust to overfitting.
Moreover, we develop an interpolation method that can enhance the validity of generated synthetic data by considering the original cluster distribution and data noise.
Finally, we conduct experiments on five realistic datasets and three classifiers, and the experimental results demonstrate the effectiveness of the proposed technique in terms of fairness and utility.
\end{abstract}




\end{frontmatter}


\begin{figure}[ht]
 \subfigure[Example: Imbalanced data]{\label{fig:data}
  \includegraphics[keepaspectratio, scale=0.8]{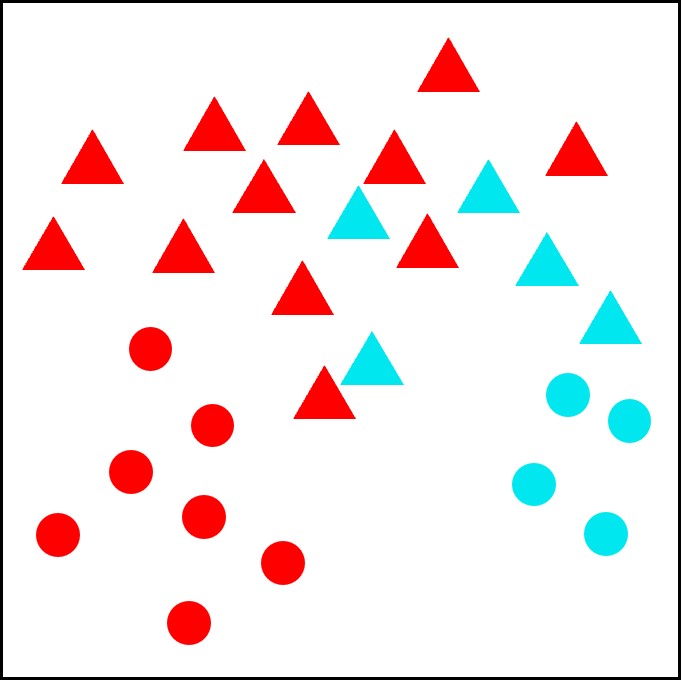}
 }
 \hspace{1cm}
 \begin{minipage}[b]{0.45\linewidth}
 \subfigure[Example: Results of the existing techniques]{\label{fig:exist}
  \centering
  \includegraphics[keepaspectratio, scale=0.8]{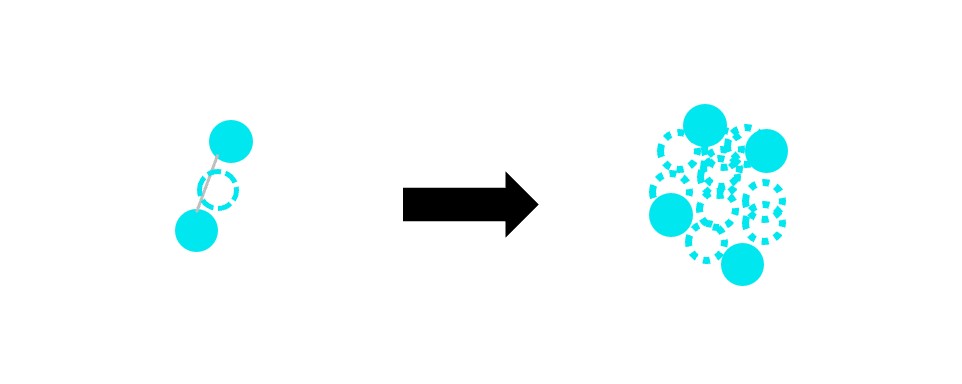}
 }
 \subfigure[Example: Results of our technique]{\label{fig:ours}
  \centering
  \includegraphics[keepaspectratio, scale=0.8]{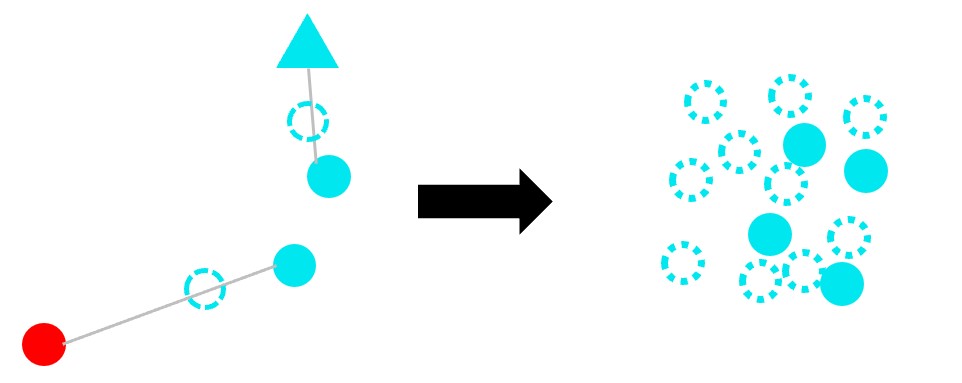}
 }
 \end{minipage}
  \caption{A toy example of fair oversampling techniques for correcting class imbalance and group imbalance in data. The triangles and circles represent a majority class and a minority class, respectively.
  Red and blue represent the majority group and minority group, respectively.
  (a) Four clusters can be identified, with the red triangle cluster having the most data and the blue circle cluster having the least data. 
  (b) Example of the synthetic data generated by existing fair oversampling techniques, which are based on homogeneous clusters, resulting in a lack of diversity in synthetic data. Such synthetic data can cause the overfitting of ML classifiers.
  (c) Example of the synthetic data generated by the proposed fair oversampling technique, which is based on heterogeneous clusters; thus, ensuring diversity in synthetic data. Such synthetic data can improve the generalization performance of ML classifiers.
}
\end{figure}
\section{Introduction}\label{sec:intro}
Machine learning (ML) classifiers have been widely used in many fields, as they improve the utility of various decision-making systems, such as loan screening, disease diagnosis, and recidivism prediction~\cite{kotsiantis2007supervised}.
Due to the increased impact of such systems on people's lives, there are concerns regarding the fairness of the decisions based on classifiers~\cite{mehrabi2021survey}.
Without proper intervention, classifiers can unintentionally discriminate against groups (e.g., race and gender); thus, worsening their fairness~\cite{DBLP:conf/nips/HardtPNS16}.
Therefore, improving the trade-off between fairness and utility is important for incorporating ML classifiers into decision-making systems.


Data imbalance is a major factor in causing the aforementioned trade-off, and it often occurs in ML pipelines.
Group imbalance in data can cause the learning of classifiers to be biased toward the majority group (e.g., male) and against the minority group (e.g., female)~\cite{DBLP:conf/nips/BolukbasiCZSK16,mehrabi2021survey}.
Moreover, learned classifiers consequently have different prediction accuracies for different groups, and thus their fairness can be deteriorated.
Many studies have addressed this issue and have aimed improving the trade-off between fairness and overall classification accuracy~\cite{DBLP:conf/cikm/IosifidisN19,DBLP:conf/icml/AgarwalBD0W18,DBLP:journals/corr/abs-1802-03765,DBLP:conf/innovations/DworkHPRZ12,DBLP:journals/kais/KamiranC11,DBLP:conf/sigsoft/ChakrabortyMM21,DBLP:conf/cikm/YanKF20,DBLP:journals/access/SalazarSAA21}.
Class imbalance is also a common problem, and it occurs in situations where data collection is difficult, such as rejected applicants in loan screening or certain diseases in disease diagnosis.
In addition, learning classifiers based on class imbalanced data is a challenging task, as the accuracy of the classifiers for minority classes is significantly deteriorated~\cite{japkowicz2002class}.
To overcome this challenge, existing studies have proposed various class balancing techniques~\cite{chawla2002smote,han2005borderline, he2008adasyn, douzas2018improving,zhou2005training,sun2007cost,freund1996experiments,freund1997decision}.
Both group and class imbalance are frequent problems in ML; however, only a few studies addressed them simultaneously.

In this work, we aim to address both group imbalance and class imbalance to improve the trade-off between fairness and utility in classifiers.
To this end, we develop a technique for balancing cluster imbalances, where each cluster is composed of a class and a group (See Figure~\ref{fig:data} that shows a toy example of cluster imbalance).

Some works have proposed balancing techniques for cluster imbalance to improve the abovementioned trade-off~\cite{DBLP:conf/sigsoft/ChakrabortyMM21,DBLP:conf/cikm/YanKF20,DBLP:journals/access/SalazarSAA21,DBLP:journals/corr/abs-2105-06345,DBLP:conf/cikm/IosifidisN19}.
Among them, fair oversampling techniques~\cite{DBLP:conf/sigsoft/ChakrabortyMM21,DBLP:conf/cikm/YanKF20,DBLP:journals/access/SalazarSAA21} are promising as they are model agnostic and can provide new useful information for classifier learning.
The goal of these techniques is to generate diverse synthetic data based on the data of minority clusters.
Moreover, these techniques have been empirically shown to achieve a good trade-off regardless of the classifier algorithm~\cite{DBLP:conf/sigsoft/ChakrabortyMM21,DBLP:conf/cikm/YanKF20,DBLP:journals/access/SalazarSAA21}.
However, existing techniques are based on homogeneous clusters, making it difficult to generate diverse data when the size of observed clusters is small.
Figure~\ref{fig:exist} illustrates this phenomenon, showing that a cluster composed of a minority class and a minority group has the smallest number of observed data and that the resulting cluster loses diversity using existing techniques.
In this case, the generalization performance of trained classifiers is not improved, resulting in overfitting.

In this paper, we propose a fair oversampling technique using heterogeneous clusters to improve the trade-off between fairness and utility in classifiers.
Unlike existing techniques, our technique can generate robust synthetic data for the abovementioned overfitting problem by considering heterogeneous clusters.
The algorithm of the proposed technique is based on an interpolation between data from a cluster and data from its heterogeneous clusters.
Overall, the contributions of this work are threefold, as mentioned below.
\begin{enumerate}
    \item To generate synthetic data using data from heterogeneous clusters, we propose a sampling method to select two types of data pairs: intra-class pairs and intra-group pairs.
    While intra-class pairs are used to generate synthetic data with features from different groups, intra-group pairs are used to generate synthetic data with features from different classes (See Figure~\ref{fig:ours} for an  illustrative example).
    These two types of pairs allow the generation of robust synthetic data that can prevent the overfitting problem of classifiers.
    \item We present an interpolation method to generate valid synthetic data from pairs of data from heterogeneous clusters.
    The proposed interpolation method can avoid generating noisy data that is far from an original cluster while ensuring diversity in synthetic data.
    To achieve this interpolation method, we consider the distance between the data and its k-nearest neighbor points and their classes.
    \item We have conducted experiments with a wide range of real-world datasets and multiple classifiers.
    The experimental results demonstrate that the proposed technique outperforms the existing ones in terms of the trade-off between fairness and utility of different measures.
\end{enumerate}

The rest of this paper is organized as follows. 
We describe works related to our study in Section~\ref{sec:related} and describe the problem formulation and definition of fairness metrics in Section~\ref{sec:preliminary}. 
Section~\ref{sec:method} shows a presentation of the proposed technique, which generates synthetic data to improve utility and fairness, and Section~\ref{sec:experiment} shows the experimental settings and results.
The conclusions are mentioned in Section~\ref{sec:conclusion}.

\section{Related Works}\label{sec:related}

\subsection*{Fairness in ML Classifiers}
Fairness has become a focus of attention in the ML classifiers integrated into decision-making systems.
There are two definitions of fairness: individual fairness~\cite{DBLP:conf/innovations/DworkHPRZ12} and group fairness~\cite{DBLP:conf/innovations/DworkHPRZ12,DBLP:conf/nips/HardtPNS16}.
Individual fairness~\cite{DBLP:conf/innovations/DworkHPRZ12} aims to ensure that similar individuals should receive similar outcomes from classifiers, while group fairness~\cite{DBLP:conf/innovations/DworkHPRZ12,DBLP:conf/nips/HardtPNS16} considers parity over different groups based on sensitive attributes (e.g., sex or race).
While individual fairness is not practical because of the difficulty in defining an appropriate similarity measure~\cite{DBLP:journals/cacm/ChouldechovaR20}, there are several mathematical definitions for group fairness that are often used in real-world regulations: statistical parity~\cite{DBLP:conf/innovations/DworkHPRZ12}, equal opportunity~\cite{DBLP:conf/nips/HardtPNS16}, and equalized odds~\cite{DBLP:conf/nips/HardtPNS16}.
Thus, group fairness has more attention in fair ML.

To account for the fairness and utility of classifiers, many studies have proposed different approaches, including pre-processing~\cite{DBLP:journals/corr/abs-1802-03765,DBLP:conf/innovations/DworkHPRZ12,DBLP:journals/kais/KamiranC11,DBLP:conf/sigsoft/ChakrabortyMM21,DBLP:conf/cikm/YanKF20,DBLP:journals/access/SalazarSAA21}, in-processing~\cite{DBLP:conf/cikm/IosifidisN19,DBLP:conf/icml/AgarwalBD0W18,du2021fairness}, and post-processing~\cite{DOHERTY2012308,feldman2015computational,DBLP:conf/nips/HardtPNS16}.
Pre-processing approaches aim to ensure fairness at the data level by mitigating potential biases in training data.
In-processing approaches attempt to incorporate constraints into learning algorithms to improve fairness.
Post-processing approaches manipulate the predictions of classifiers to correct unfairness in predictions.
Among various approaches, pre-processing approaches have become popular due to their classifier agnostic nature.

One of the main challenges of the abovementioned existing pre-processing approaches is that they consider the overall classification accuracy and thus cannot address the class imbalance problem in training data.
In this paper, we aim to develop a pre-processing approach that addresses this challenge.

\subsection*{Class Imbalance in ML}
Class imbalance often occurs in the cases in which it is difficult to collect data for a particular class (e.g., credit risk screening, case diagnosis, and spam determination).
Thus, the distribution of classes in training data is often not very balanced.
ML classifiers trained on such data may have worse accuracy for minority classes, as the algorithms of classifiers are constructed based on the theory of empirical risk minimization and tend to favor majority classes~\cite{DBLP:journals/tkde/HeG09}.

To address the class imbalance problem, many studies have proposed different approaches: data level~\cite{chawla2002smote,han2005borderline, he2008adasyn, douzas2018improving}, algorithmic level~\cite{zhou2005training,sun2007cost}, and hybrid learning~\cite{freund1996experiments,freund1997decision}.
Most existing studies focus on data-level approaches, as they can be applied regardless of the used ML classifiers.
Data level approaches can be further divided into two techniques: undersampling and oversampling.
Oversampling techniques are widely used because of concerns that the undersampling techniques can worsen classifier performance due to the loss of a large amount of training data.
The simplest oversampling technique is to duplicate the instances of minority classes for balancing class distribution.
However, this technique can cause overfitting of ML classifiers due to the lack of diversity in generated data.

To tackle this problem, the Synthetic Minority Oversampling Technique (SMOTE) has been proposed~\cite{chawla2002smote}.
SMOTE can generate synthetic data that can bring new information to learning classifiers.
The algorithm of SMOTE is simple: (1) randomly select an instance from the minority class, (2) randomly select an instance from the K nearest neighborhoods of that instance, (3) and generate a synthetic instance based on the interpolation between the two instances.
Due to the great success of SMOTE against class imbalances, many extensions of SMOTE have been proposed to improve its performance under different scenarios~\cite{han2005borderline, he2008adasyn, douzas2018improving,bunkhumpornpat2009safe,abdi2015combat,feng2021investigation}, such as techniques that generate synthetic data near the class boundary~\cite{han2005borderline,he2008adasyn}, techniques that do not generate synthetic data according to the class density of the neighborhoods~\cite{bunkhumpornpat2009safe,abdi2015combat}, and techniques that generate data in sparse regions to avoid duplication~\cite{douzas2018improving,feng2021investigation}.

While the above techniques aim to improve utility, we aim to have a technique that can improve the trade-off between fairness and utility.
Thus, we develop an oversampling technique that can correct the class imbalance and group imbalance based on SMOTE.

\section{Preliminaries}\label{sec:preliminary}
We consider the setting in which one predicts the class label $y \in Y$ from the $D$-dimensional feature vector $x\in \mathbb{R}^D$ based on the training dataset $\mathcal{T} = \{ (x^i, y^i) \}^{N}_{i=1}$, where $N$ is the number of observed instances.
In this study, the class label is assumed to be binary $Y \in \{0,1\}$.
We also assume that each instance is assigned to exactly one group $g \in G$ based on the protected attributes (e.g., race and gender) included in its feature vector.
We consider $M$ categorical groups $G\in [M]$, where groups are prespecified by ML practitioners.
For example, ML practitioners can consider groups based on race (e.g., white and black) or intersectional groups of race and gender (e.g., white women, white men, black women, and black men).

The goal of fair ML is to minimize disparity (i.e., unfairness) while maintaining utility with respect to the prediction $\hat{Y}$ of $Y$.
There are two metrics to assess disparity: individual fairness and group fairness.
Since the latter is taken into account in many regulations and existing studies, we also consider only group fairness for disparity.
The disparity based on group fairness metric $F$ is defined as follows:
\begin{equation}\label{eq:dis}
  | \max_{g \in G} F_g - \min_{g \in G} F_g |, 
\end{equation}
where $F_g$ represents a group fairness metric for a group $g$.

We introduce three group fairness metrics that are used in many studies: statistical parity~\cite{DBLP:conf/innovations/DworkHPRZ12}, equal opportunity~\cite{DBLP:conf/nips/HardtPNS16}, and equalized odds~\cite{DBLP:conf/nips/HardtPNS16}.
\begin{defn}
 Statistical parity measures the difference between the positive prediction rate for a group $g$ and that of all instances.
 \begin{equation}\label{eq:demo}
     F_g = P(\hat{Y}=1) - P(\hat{Y}=1\mid G=g).
 \end{equation}
\end{defn}

\begin{defn}
 Equal opportunity measures the difference between the true positive rate for a group $g$ and that of all instances.
 \begin{equation}\label{eq:eoo}
     F_g = P(\hat{Y}=1 \mid Y=1) - P(\hat{Y}=1\mid Y=1, G=g).
 \end{equation}
\end{defn}

\begin{defn}
 Equalized odds measures the difference between the true positive and false positive rates for a group $g$ and those of all instances.
 \begin{equation}\label{eq:eod}
     F_g = P(\hat{Y}=y \mid Y=y) - P(\hat{Y}=y\mid Y=y, G=g).
 \end{equation}
\end{defn}

All the above fairness metrics are related to the outputs of trained classifiers.
However, data imbalances are known to have a significant impact on fairness.
Therefore, it is important to quantitatively assess data imbalances.
Thus, we introduce a statistical metric to investigate data imbalances.
Concerning fairness, not only the imbalance between classes but also that between groups are considered to be important.
Therefore, we evaluate imbalances on a cluster basis.
Each cluster $C_{y,g}$ is the intersection of class $y$ and group $g$, i.e., $C_{y,g}=\{i\mid y^i = y, g^i = g \}$.

Then the degree of imbalance of each cluster is evaluated by the difference between its size and the largest cluster among all clusters.
\begin{equation}\label{eq: imbalance}
    \max_{y\in Y, g\in G}{|C_{y,g}|} - |C_{y,g}|
\end{equation}
The larger the difference in the size of clusters, the more biased a classifier's predictions can be. 
We believe that this fact affects the trade-off between fairness and utility of the classifiers.
In the next section, we describe how to correct this imbalance by applying the proposed oversampling technique to improve fairness and utility.

\section{Method}\label{sec:method}
In this section, we present the newly developed fair oversampling technique to address fairness in class imbalanced data.
The main working procedure involves the following three steps:
(1) Randomly select an instance $i$ from a cluster $C_{y,g}$ that should be oversampled.
(2) Randomly select an instance $j$ from the heterogeneous clusters of $C_{y,g}$ using our sampling method.
(3) Generate a synthetic instance by interpolating $x^i$ and $x^j$ with our interpolation method.
The details of Step (2) and Step (3) are described in Section~\ref{subsec:sshc} and Section~\ref{subsec:interpolation}, respectively.

Before describing the details of each step, we first discuss the state-of-the-art algorithm of the fair oversampling technique, Fair SMOTE (FSMOTE)~\cite{DBLP:conf/sigsoft/ChakrabortyMM21}.
FSMOTE is an extension of the SMOTE algorithm~\cite{chawla2002smote}, and its goal is to address the cluster imbalances Eq.~\eqref{eq: imbalance}. 
The main working steps of the FSMOTE algorithm are as follows:
(1) Randomly select an instance $i$ from a cluster $C_{y,g}$ that should be oversampled.
(2) Randomly select an instance $j$ from the set of K nearest neighbors (KNN) of that instance $i$ that belongs to the same cluster. 
(3) Generate a synthetic instance by interpolating $x^i$ and $x^j$.
The formula for generating a new synthetic instance $(x^{new}, y^{new})$ is as follows:
\begin{equation}\label{eq:smote}
    x^{new} = x^i + w \times (x^j - x^i),\ \ \ y^{new} = y^i
\end{equation}
where $w$ is the interpolation weight drawn from the uniform distribution $U(0, 1)$.

As seen from the above steps, FSMOTE generates synthetic instances based on homogeneous clusters. 
Therefore, if the cluster is too small (e.g., a cluster of a minority class belongs to a minority group), FSMOTE cannot generate diverse synthetic data, and the synthetic data may worsen the performance of classifiers.
The proposed oversampling technique can avoid this problem by considering data from heterogeneous clusters.
Therefore, the synthetic data generated by the proposed technique is diverse and can provide new information useful for learning classifiers.
The following sections describe the details of the proposed algorithm.

\subsection{Sampling Instance Pairs from Heterogeneous Clusters}\label{subsec:sshc}
In this section, we explain our sampling method to select instance pairs for generation from heterogeneous clusters.
We first define the heterogeneous clusters.

We consider two heterogeneous clusters $H_y$ and $H_g$.
Given a target cluster $C_{y,g}$ that should be oversampled, $H_y$ is the cluster that has the different class and same group (i.e.,  $H_y = \{j\mid y^j\neq y, g^j = g \}$), and $H_g$ is the cluster that has the same class and different group (i.e.,  $H_g = \{j\mid y^j = y, g^j\neq g \}$).
We use the cluster $H_y$ for sampling intra-group pairs and the cluster $H_g$ for sampling intra-class pairs.

The intra-group instance pairs are used to generate synthetic instances that have class-mix features.
The class-mix features can mitigate the spurious correlation between classes and features~\cite{DBLP:conf/iclr/ZhangCDL18}, allowing the classifier to learn the class boundary more correctly and improve the prediction accuracy of each class in a given group.
The intra-class instance pairs are used to generate synthetic instances that have group-mix features.
The group-mix features can mitigate the correlation between a certain class and protected attributes~\cite{du2021fairness}, thus promoting the independence of classifier predictions and protected attributes and improving group fairness.
Thus, synthetic data with these features can improve the generalization performance of classifiers.

Our sampling method randomly selects intra-group or intra-class instance pairs for a target cluster $C_{y,g}$ with the probability $p_{y,g}$ and $1-p_{y,g}$.
We define the probability $p_{y,g}$ as follows:
\begin{equation}\label{eq:select}
    p_{y,g}=\frac{|H_y|}{|H_y| + |H_g|}.
\end{equation}
According to the Bernoulli distribution with $p_{y,g}$ (i.e., $\text{Bernoulli}(p_{y,g})$), we select which pairs to use each time we generate new synthetic data.
This allows us to sample more data from heterogeneous clusters with a larger number of observed data.

\subsection{Interpolation between the Heterogeneous Instances}\label{subsec:interpolation}
In this section, we describe an interpolation method between two instances belonging to heterogeneous clusters.
Using the proposed interpolation method, we can generate a valid synthetic instance using the instance $i$ of the target cluster $C_{y,g}$ and the instance $j$ of the heterogeneous cluster $H_y$ or $H_g$.
In general, generating a valid synthetic instance by the existing interpolation method Eq.~\eqref{eq:smote} is difficult, as the distance between $x^i$ and $x^j$ is large.
To address this issue, we normalize the features of the generated synthetic instance using the distance between $x^i$ and its K nearest neighbors KNN$(i)$.

We propose a data interpolation method for a instance pair from heterogeneous clusters.
\begin{equation}\label{eq:interpolation}
    x^{new} = x^i + w^i \times (x^j - x^i) \times \frac{\max_k d(i, \mathrm{KNN}(i))}{d(i, j)}, 
\end{equation}
where $d(i,j)$ is the distance between $x^i$ and $x^j$, and $w^i$ is the interpolation weight for $x^i$. 
We use the euclidean distance as $d$ in this paper.
In the second term on the RHS of Eq.\eqref{eq:interpolation}, we normalize the vector distance between $x^i$ and $x^j$ and multiply it by the maximum distance between $x^i$ and the instance of KNN$(i)$.
Thus, the generated synthetic data can follow the original distribution of the original cluster and maintain validity.
Moreover, the distance to the most distant neighbors can ensure more diversity in the synthetic data~\cite{feng2021investigation}.
Hence, we can generate synthetic instances that are robust to overfitting and that can preserve the original distribution of the cluster.

We also propose a different distribution of the interpolation weight from the existing one, which is generated from the uniform distribution $U(0, 1)$. 
However, the proposed technique is based on heterogeneous clusters and is prone to overlapping between classes.
Since overlapping between classes has been shown to reduce classifier performance, avoiding this phenomenon is important for oversampling~\cite{han2005borderline,bunkhumpornpat2009safe}.
Thus, we use a KNN-based local density that evaluates the overlap ratio of each instance.
The local density $\Delta^i$ of an instance $i$ is the fraction of its KNN that is different from the class $y^i$.
\begin{equation}\label{eq:kdn}
    \Delta^i = \frac{ | \{k\mid x^k \in \mathrm{KNN}(x^i), y^k = y^i \} | }{K}.
\end{equation}
Intuitively, if $\Delta^i$ is small (large), an instance $i$ is likely (not) to be overlapped.
Using this $\Delta^i$, we denote the distribution $U(0, \Delta^i)$ that $w^i$ follows, allowing us to avoid the overlapping of the data generated based on the interpolation with heterogeneous clusters.

The summary of our algorithm is shown in Algorithm~\ref{al:all}.

\begin{algorithm}[ht]
\caption{Proposed Fair Oversampling Algorithm}\label{al:all}
\begin{algorithmic}
\REQUIRE Training dataset $\mathcal{T}$, the number of nearest neighbors $K$
\STATE composes clusters from the training dataset $\{C_{y,s}\mid y\in Y, \in G\}$.
\STATE calculate maximum cluster size $S=\max_{y\in Y, g\in G}{|C_{y,g}|}$.
\STATE create an empty set to add synthetic data $\mathcal{T}^{new}$.
\FOR{$C_{y,g} \in \mathcal{T}$} 
\STATE compose heterogeneous clusters $H_y = \{j\mid y^j\neq y, g^j = g \}$ and $H_g = \{j\mid y^j = y, g^j\neq g \}$.

\FOR{$s=1,\dots, S-|C_{y,g}|$}
 \STATE randomly sample  $x^i$ from $C_{y,g}$.
 \STATE calculate $p_{y,g}$ using Eq.~\eqref{eq:select}.
 \STATE draw $b\sim \text{Bernoulli}(p_{y,g})$.
 \IF{$b$ is True}
  \STATE randomly sample $x^j$ from $H_y$.
 \ELSE 
  \STATE randomly sample $x^j$ from $H_g$.
  \ENDIF
 \STATE generate synthetic data $x^{new}$ using Eq.~\eqref{eq:interpolation} with $(x^i,x^j)$.
 \STATE $\mathcal{T}^{new} = \mathcal{T}^{new} \cup (x^{new}, y)$
\ENDFOR

\ENDFOR
\RETURN $\mathcal{T} \cup \mathcal{T}^{new}$
\end{algorithmic}
\end{algorithm}

\section{Experiments}\label{sec:experiment}

\subsection{Datasets}\label{subsec:datasets}
In this work, we adopt five datasets that have been widely investigated in previous fairness studies, and their characteristics are summarized in Table~\ref{tab:dataset}.
\begin{table}[t]
    \centering
    \caption{Description of the $5$ benchmark datasets with their characteristics, 
    including the number of instances $N$, dimensions of the feature vector $D$, number of groups $M$, class distribution, and group distribution.}
    \label{tab:dataset}
    \begin{tabular}{cccccc} \hline
        Dataset & $N$ & $D$ & $M$ & Class Distribution & Group Distribution\\ \hline
        Adult & $48842$ & $12$ & $4$ & $37155/11687$ & $3915/28735/3165/13027$ \\
        German  & $1000$ & $30$ & $4$ & $300/700$ & $490/139/200/171$ \\
        Heart  & $297$ & $18$ & $2$ & $160/137$ & $201/96$ \\
        COMPAS   & $7214$ & $14$ & $4$ & $3251/3963$ & $3932/828/1887/567$ \\
        C$\&$C   & $1994$ & $146$ & $2$ & $1386/608$ & $1385/609$ \\ \hline
    \end{tabular}
\end{table}

The first dataset is the Adult dataset~\cite{Dua:2019}.
The task is to predict whether a person’s income is more than $50k$ per year. 
We use four groups based on the intersection of gender (male/female) and race (black/white).
The second dataset is the German Credit dataset~\cite{Dua:2019}.
The task is to predict whether an individual is a good or bad credit risk. 
We form four groups based on the intersection of gender (male/female) and age (over $30$/under $30$).
The third is the Heart Disease dataset~\cite{janosi1988heart}.
The task is to predict whether a patient has heart disease. 
We use two groups based on gender (male/female).
The fourth dataset is the COMPAS dataset~\cite{propublica2018}.
The task is to predict recidivism based on a defendant's risk score.
We use four groups based on the intersection of gender (male/female) and race (black/white).
The fifth dataset is the Community and Crime dataset~\cite{Dua:2019}.
The task is to predict whether a community has a high (above the $70$ percentile) crime rate.
We form two groups based on the percentage of Black (higher/lower than the median).

\subsection{Experimental Settings}\label{subsec:setting}
To verify the effectiveness of the proposed technique, we select $5$ techniques for comparison purposes.
Two of these techniques (FBSMOTE and FADASYN) are newly developed in this study. 
FBSMOTE and FADASYN are simple extensions of existing techniques (Borderline-SMOTE~\cite{han2005borderline} and ADASYN~\cite{he2008adasyn}). 
However, they may outperform the state-of-the-art technique, FSMOTE, because they consider heterogeneous clusters in their data generation.
We consider these techniques to confirm the effectiveness of our technique in depth.
The summary of the comparison techniques is as follows:
\begin{itemize}
    \item Original: this technique does not modify the imbalances and uses the original training dataset as is.
    \item SMOTE\footnote{\url{https://imbalanced-learn.org/stable/references/generated/imblearn.over_sampling.SMOTE.html}}~\cite{chawla2002smote}:
    this technique is the most widely used oversampling technique for correcting class imbalance.
    \item FSMOTE\footnote{\url{https://github.com/joymallyac/Fair-SMOTE}}~\cite{DBLP:conf/sigsoft/ChakrabortyMM21}: this technique is the state-of-the-art fair oversampling technique for correcting the cluster imbalance Eq.~\eqref{eq: imbalance}.
    \item FBSMOTE: we adapt Borderline-SMOTE~\cite{han2005borderline} to balance the cluster imbalance Eq.~\eqref{eq: imbalance}, called Fair Borderline-SMOTE (FBSMOTE). 
    This technique was developed in the scope of this work. 
    The implementation of FBSMOTE applies Borderline-SMOTE\footnote{\url{https://imbalanced-learn.org/stable/references/generated/imblearn.over_sampling.BorderlineSMOTE.html}} to each cluster.
    \item FADASYN: we adapt ADASYN~\cite{he2008adasyn} to balance the cluster imbalance Eq.~\eqref{eq: imbalance}, called Fair ADASYN (FADASYN). 
    This technique was developed in the scope of this work. 
    
    The implementation of FADSYN applies ADASYN\footnote{\url{http://glemaitre.github.io/imbalanced-learn/auto_examples/over-sampling/plot_adasyn.html}} to each cluster.
\end{itemize}
The nearest neighbor number $K$ of all the techniques in our experiment is fixed at $5$.
The optimal imbalance ratio is not obvious and its exploration is outside the scope of our study.
Therefore, we fix the parameters related to the imbalance ratio set in each technique.
Also, for a fair comparison, we will not implement the method of removing noisy instances after oversampling as proposed in\cite{DBLP:conf/sigsoft/ChakrabortyMM21}.

In all our experiments, we use three different classifiers: Logistic Regression (LR), Support Vector Machine (SVM), and Naive Bayes (NB).
All classifiers are used as implemented in the Python library scikit-learn~\cite{DBLP:journals/jmlr/PedregosaVGMTGBPWDVPCBPD11} with default parameters.

\begin{table}[t]
\centering
 \caption{Confusion matrix.}
 \label{tab:cm}
  \begin{tabular}{cc|cc} \hline
   & & \multicolumn{2}{c}{Predicted}\\ 
    & & Positive & Negative\\ \hline
    \multirow{2}{*}{Actual}
    & Positive     & TP & FN \\
    & Negative     & FP & TN  \\ \hline
  \end{tabular}
\end{table}
We focus on the fairness and utility metrics in all experiments.
For fairness, we use the disparity Eq.~\eqref{eq:dis} based on the three group fairness metrics defined in~\Cref{eq:demo,eq:eoo,eq:eod}: statistical parity (SP), equal opportunity  (E. Opp.), and equalized odds (E. Odds).
For utility, we use the balanced accuracy score (BAcc), which is widely used in imbalanced classification problems, as the accuracy score (the fraction of right predictions) is no longer considered to be an excellent utility metric.
BAcc is the mean of the true positive rate and true negative rate.
\begin{equation}\label{eq:bacc}
    BAcc = \frac{1}{2} \left( \frac{TP}{TP+FN} + \frac{TN}{TN+FP} \right),
\end{equation}
where TP, TN, FP, and FN are defined in Table~\ref{tab:cm}.

To evaluate the proposed technique, we use 5-fold cross-validation for all experiments.
Each metric is reported in the form of an average.
The hyperparameters of the classifiers and oversampling techniques are fixed for all runs.

\subsection{Results}
\Cref{tab:lr_result,tab:svm_result,tab:nb_result} show the performance of 6 comparative techniques in terms of 5 datasets using the LR, SVM, and NB classifiers, respectively. 
Based on the results shown in these three tables, we can obtain the following conclusions.

(1) The fair oversampling techniques (FSMOTE, FBSMOTE, FADASYN, and Ours) can improve all three fairness metrics well compared with the other techniques. 
This fact demonstrates that fair oversampling techniques which address group and class imbalances improve fairness.

(2) The BAcc. of the oversampling techniques is better than the original technique (i.e., a technique without any modification of data imbalance).
This fact demonstrates that correcting for class imbalance by generating synthetic data is effective with regard to improving the accuracy of classifier predictions for each class.

(3) The fairness of SMOTE is severely compromised.
This fact is because SMOTE exacerbated group imbalance by only addressing class imbalance.
That is, SMOTE ignores minority groups, as it randomly selects instances within the minority class.

(4) The three fair oversampling techniques (FSMOTE, FBSMOTE, and FADASYN) do not perform as the proposed technique.
These techniques generate synthetic data based on homogeneous clusters; thus, limiting their generation area to the inside of the observed cluster regions.
Thus, they cannot generate effective samples when the observed clusters are small.
This phenomenon results in the overfitting of classifiers and worsens their performances.

(5) The proposed technique is fairer than other techniques and can preserve balanced accuracy (BAcc.).
The results of the proposed technique are stable regardless of the datasets or classifiers, which demonstrates the effectiveness of the proposed technique for generating data by considering heterogeneous clusters.
FBSMOTE and FADASYN also consider heterogeneous clusters and sometimes achieve the best fairness.
However, our proposed technique that is based on a new interpolation method can often produce better results than FBSMOTE and FADASYN, because it can generate more diverse and valid synthetic data.

\begin{table}[ht!]
 \caption{Experiment results using the LR classifier. The technique achieving the lowest (best) fairness or highest (best) utility is shown in bold.}
 \label{tab:lr_result}
  \begin{tabular}{c|c|c|ccc} \hline
   Dataset& Technique & Utility& \multicolumn{3}{c}{Fairness}\\ 
    &  & BAcc & SP & E. Opp. & E. Odds\\ \hline \hline
    \multirow{5}{*}{Adult}
    & Original  & 0.7631           & \textbf{0.222}  & 0.1848           & 0.1365 \\
    & SMOTE     & \textbf{0.8208}  & 0.4118          & 0.2727           & 0.2739 \\
    & FSMOTE    & 0.8124           & 0.2754          & 0.1439           & 0.1081  \\ 
    & FBSMOTE   & 0.7769           & 0.3315          & 0.2676           & 0.2321  \\
    & FADASYN   & 0.8018           & 0.3458          & 0.1957           & 0.2043 \\
    & Ours      & 0.8125           & 0.261           & \textbf{0.0642}  & \textbf{0.0678}  \\ \hline \hline
    \multirow{5}{*}{German Credit}
    & Original  &  0.624           & 0.265           & 0.2449           & 0.2886   \\
    & SMOTE     &  \textbf{0.6936} & 0.2707          & 0.2855           & 0.2407  \\
    & FSMOTE    &  0.6891          & 0.1943          & 0.236            & 0.2008  \\ 
    & FBSMOTE   &  0.6671          & 0.2072          & 0.2729           & 0.2139 \\
    & FADASYN   &  0.68            & 0.1882          & 0.2275           & 0.2272 \\
    & Ours      &  0.6883          &\textbf{0.1265}  & \textbf{0.1581}  & \textbf{0.1508}   \\ \hline \hline
    \multirow{5}{*}{Heart}
    & Original  &  0.8235          & 0.3229          & 0.2116           & 0.1304 \\
    & SMOTE     &  0.8221          & 0.3557          & 0.2206           & 0.1458 \\
    & FSMOTE    &  \textbf{0.8278} & 0.2675          & 0.2268           & 0.1403 \\
    & FBSMOTE   &  0.8227          & 0.2507          & 0.2341           & 0.1249 \\
    & FADASYN   &  0.8137          & \textbf{0.2319} & 0.2519           & 0.1293 \\
    & Ours      &  0.822           & 0.2369          & \textbf{0.1572}  & \textbf{0.09648}    \\ \hline \hline
    \multirow{5}{*}{COMPAS}
    & Original  &  0.6537          & 0.3983          & 0.2806           & 0.377  \\
    & SMOTE     &  \textbf{0.6626} & 0.3763          & 0.264            & 0.3527  \\
    & FSMOTE    &  0.6574          & 0.1267          & 0.1519           & 0.117 \\
    & FBSMOTE   &  0.5884          & 0.1774          & 0.2039           & 0.1804  \\
    & FADASYN   &  0.6506          & 0.1232          & \textbf{0.088}   & 0.1026 \\
    & Ours      &  0.6551          & \textbf{0.1064} & 0.1301           & \textbf{0.1012} \\ \hline \hline
    \multirow{5}{*}{C$\&$C}
    & Original  &  0.8221          & 0.5854          & 0.3956            & 0.3587 \\
    & SMOTE     &  \textbf{0.8477} & 0.6509          & 0.3167            & 0.3974 \\
    & FSMOTE    &  0.797           & 0.3339          & 0.06308           & 0.05502  \\
    & FBSMOTE   &  0.7872          & 0.3254          & 0.03998           & 0.04942  \\
    & FADASYN   &  0.7845          & 0.299           & 0.08312           & 0.05936 \\
    & Ours      &  0.7698          & \textbf{0.2839}  & \textbf{0.0003}  & \textbf{0.0071}   \\ \hline \hline
  \end{tabular}
\end{table}

\begin{table}[ht!]
 \caption{Experiment results using the SVM classifier. The technique achieving the lowest (best) fairness or highest (best) utility is shown in bold.}
 \label{tab:svm_result}
  \begin{tabular}{c|c|c|ccc} \hline
   Dataset& Technique &  Utility& \multicolumn{3}{c}{Fairness}\\ 
    &  & BAcc & SP & E. Opp. & E. Odds\\ \hline \hline
    \multirow{5}{*}{Adult}
    & Original   & 0.7402           & \textbf{0.2141}  & 0.2174           & 0.1525 \\
    & SMOTE      & 0.8081           & 0.4464           & 0.3957           & 0.3526 \\
    & FSMOTE     & 0.8084           & 0.3478           & 0.2878           & 0.245  \\ 
    & FBSMOTE    & 0.7817           & 0.4398           & 0.4687           & 0.3762  \\
    & FADASYN    & 0.8045           & 0.441            & 0.4694           & 0.3809  \\
    & Ours       & \textbf{0.8108}  & 0.2959           & \textbf{0.1127}  & \textbf{0.1369}   \\ \hline \hline
    \multirow{5}{*}{German Credit}
    & Original  & 0.6172           & 0.2555            & 0.2157           & 0.2901 \\
    & SMOTE     & 0.6681           & 0.2098            & 0.2215           & 0.1807 \\
    & FSMOTE    & 0.6631           & 0.2288            & 0.1788           & 0.2549\\
    & FBSMOTE   & 0.6617           & 0.1995            & 0.2089           & 0.2248 \\
    & FADASYN   & 0.6591           & 0.2159            & \textbf{0.1778}  & 0.239 \\
    & Ours      & \textbf{0.6774}  & \textbf{0.184}    & 0.1981           & \textbf{0.1679}   \\ \hline \hline
    \multirow{5}{*}{Heart}
    & Original  & 0.801            & 0.3442            & 0.2611           & 0.1559 \\
    & SMOTE     & \textbf{0.812}   & 0.3461            & 0.2303           & 0.1511 \\
    & FSMOTE    & 0.7835           & 0.2797            & 0.2347           & 0.1448  \\
    & FBSMOTE   & 0.7824           & 0.2631            & 0.2252           & 0.1732 \\
    & FADASYN   & 0.804            & 0.3017            & 0.243            & 0.1547 \\
    & Ours      & 0.7798           & \textbf{0.208}    & \textbf{0.1857}  & \textbf{0.09634}     \\ \hline \hline
    \multirow{5}{*}{COMPAS}
    & Original  & 0.66             & 0.3436            & 0.2554            & 0.3086 \\
    & SMOTE     & 0.659            & 0.2681            & 0.1873            & 0.2467 \\
    & FSMOTE    & \textbf{0.6686}  & 0.2812            & 0.2278            & 0.2373  \\ 
    & FBSMOTE   & 0.5882           & 0.2281            & 0.3754            & 0.2224 \\
    & FADASYN   & 0.6419           & 0.2476            & 0.2791            & 0.2262 \\
    & Ours      & 0.6675           & \textbf{0.2104}   & \textbf{0.1825}   & \textbf{0.1682}   \\ \hline \hline
    \multirow{5}{*}{C$\&$C}
    & Original  & 0.8237           & 0.6509            & 0.4787            & 0.4439 \\
    & SMOTE     & \textbf{0.8479}  & 0.7049            & 0.3655            & 0.4634 \\
    & FSMOTE    & 0.8013           & 0.3445            & 0.0626            & 0.05982  \\
    & FBSMOTE   & 0.7823           & 0.3104            & \textbf{0.05512}  & \textbf{0.03622} \\
    & FADASYN   & 0.787            & 0.3162            & 0.06698           & 0.06116 \\
    & Ours      & 0.7594           & \textbf{0.2429}   & 0.1105            & 0.0593   \\ \hline \hline
  \end{tabular}
\end{table}

\begin{table}[ht!]
 \caption{Experiment results using the NB classifier. The technique achieving the lowest (best) fairness or highest (best) utility is shown in bold.}
 \label{tab:nb_result}
  \begin{tabular}{c|c|c|ccc} \hline
   Dataset& Technique &  Utility& \multicolumn{3}{c}{Fairness}\\ 
    &  & BAcc & SP & E. Opp. & E. Odds\\ \hline \hline
    \multirow{5}{*}{Adult}
    & Original   & \textbf{0.7182}  & 0.4222           & 0.1655           & 0.2555 \\
    & SMOTE      & 0.7              & 0.3829           & 0.1818           & 0.245 \\
    & FSMOTE     & 0.6885           & 0.1783           & 0.05498          & 0.07438   \\ 
    & FBSMOTE    & 0.669            & 0.1618           & 0.07282          & 0.08856 \\
    & FADASYN    & 0.6306           & 0.1243           & 0.07212          & 0.09606 \\
    & Ours       & 0.6585           & \textbf{0.1164}  & \textbf{0.0264}  & \textbf{0.0502}   \\ \hline \hline
    \multirow{5}{*}{German Credit}
    & Original   & \textbf{0.6674}  & 0.2687            & 0.2568          & 0.2433 \\
    & SMOTE      & 0.6602           & 0.2477            & 0.291           & 0.2246 \\
    & FSMOTE     & 0.6634           & 0.1552            & 0.1903          & 0.1643  \\ 
    & FBSMOTE    & 0.6469           & 0.153             & 0.22            & 0.1693 \\
    & FADASYN    & 0.6424           & 0.1449            & 0.1854          & 0.147 \\
    & Ours       & 0.6376           & \textbf{0.1232}   & \textbf{0.1441} & \textbf{0.1314}   \\ \hline \hline
    \multirow{5}{*}{Heart}
    & Original   & 0.8086           & 0.2761            & 0.1622          & 0.0831 \\
    & SMOTE      & \textbf{0.8154}  & 0.2915            & 0.1616          & 0.09466 \\
    & FSMOTE     & 0.7933           & \textbf{0.1931}   & 0.1702          & 0.09852  \\
    & FBSMOTE    & 0.796            & 0.2167            & 0.2716          & 0.1639 \\
    & FADASYN    & 0.8044           & 0.2014            & 0.2395          & 0.1331 \\
    & Ours       & 0.7944           & 0.2026            & \textbf{0.1548} & \textbf{0.08062}     \\ \hline \hline
    \multirow{5}{*}{COMPAS}
    & Original   & \textbf{0.6241}  & 0.62              & 0.5289          & 0.6121 \\
    & SMOTE      & 0.6137           & 0.6806            & 0.6147          & 0.669 \\
    & FSMOTE     & 0.61             & 0.1491            & 0.1501          & 0.13278  \\ 
    & FBSMOTE    & 0.5705           & 0.1456            & 0.1974          & 0.1429 \\
    & FADASYN    & 0.6004           & 0.1262            & \textbf{0.1159} & 0.1114 \\
    & Ours       & 0.6161           & \textbf{0.088}    & 0.1728          & \textbf{0.1059}  \\ \hline \hline
    \multirow{5}{*}{C$\&$C}
    & Original   & 0.7176           & 0.2941            & 0.1172          & 0.1092 \\
    & SMOTE      & \textbf{0.7238}  & 0.3043            & 0.1159          & 0.1137 \\
    & FSMOTE     & 0.6999           & 0.2248            & 0.07886         & 0.06744   \\ 
    & FBSMOTE    & 0.7138           & 0.2249            & 0.1329          & 0.08966 \\
    & FADASYN    & 0.7244           & 0.2463            & 0.1187          & 0.07464 \\
    & Ours       & 0.6943           & \textbf{0.2138}  & \textbf{0.07538} & \textbf{0.05788}    \\ \hline \hline
  \end{tabular}
\end{table}

\section{Conclusion}\label{sec:conclusion}
Ensuring ML fairness in class imbalanced environments is an open challenge in real-world applications, and the existing fair oversampling techniques that address this challenge are based on homogeneous clusters, where each cluster is the intersection of a single group and a single class.
We could find that these techniques may result in the overfitting of ML classifiers if the observed clusters are small enough.
In this work, we developed an oversampling technique that is robust to overfitting through the use of heterogeneous clusters.
The proposed technique is based on the interpolation of instances from a cluster and its heterogeneous clusters while considering the valid coverage of the cluster.
To achieve this, we presented a sampling method for selecting instance pairs to generate diverse synthetic instances.
Moreover, we introduced a interpolation method to ensure the validity of the generated synthetic instances.
Therefore, synthetic instances generated by the proposed technique are more diverse and valid than those generated by existing techniques.
In addition, we discussed the performance of fairness oversampling techniques using five real-world datasets and basic three classifiers (Logistic Regression, Support Vector Machine, and Naive Bayes).
The experimental results showed that the proposed technique outperforms existing techniques in the trade-off between fairness and utility, and that this performance is not limited to the chosen classifier algorithm.

In future works, we plan to study how our technique can affect performance in more complex classifiers (e.g., deep learning classifiers and classifiers with fairness constraints).
We also plan to study the cases in which groups are not categorical (e.g., proportion of each race in a given area).



 \bibliographystyle{elsarticle-num} 
 \bibliography{Manuscript}

\begin{thebibliography}{10}
\expandafter\ifx\csname url\endcsname\relax
  \def\url#1{\texttt{#1}}\fi
\expandafter\ifx\csname urlprefix\endcsname\relax\def\urlprefix{URL }\fi
\expandafter\ifx\csname href\endcsname\relax
  \def\href#1#2{#2} \def\path#1{#1}\fi

\bibitem{kotsiantis2007supervised}
S.~B. Kotsiantis, I.~Zaharakis, P.~Pintelas, et~al.,
  \href{http://www.booksonline.iospress.nl/Content/View.aspx?piid=6950}{Supervised
  machine learning: A review of classification techniques}, Emerging artificial
  intelligence applications in computer engineering 160~(1) (2007) 3--24.
\newline\urlprefix\url{http://www.booksonline.iospress.nl/Content/View.aspx?piid=6950}

\bibitem{mehrabi2021survey}
N.~Mehrabi, F.~Morstatter, N.~Saxena, K.~Lerman, A.~Galstyan,
  \href{https://doi.org/10.1145/3457607}{A survey on bias and fairness in
  machine learning}, ACM Computing Surveys (CSUR) 54~(6) (2021) 1--35.
\newline\urlprefix\url{https://doi.org/10.1145/3457607}

\bibitem{DBLP:conf/nips/HardtPNS16}
M.~Hardt, E.~Price, N.~Srebro,
  \href{https://proceedings.neurips.cc/paper/2016/hash/9d2682367c3935defcb1f9e247a97c0d-Abstract.html}{Equality
  of opportunity in supervised learning}, in: Advances in Neural Information
  Processing Systems 29, 2016, pp. 3315--3323.
\newline\urlprefix\url{https://proceedings.neurips.cc/paper/2016/hash/9d2682367c3935defcb1f9e247a97c0d-Abstract.html}

\bibitem{DBLP:conf/nips/BolukbasiCZSK16}
T.~Bolukbasi, K.~Chang, J.~Y. Zou, V.~Saligrama, A.~T. Kalai,
  \href{https://proceedings.neurips.cc/paper/2016/hash/a486cd07e4ac3d270571622f4f316ec5-Abstract.html}{Man
  is to computer programmer as woman is to homemaker? debiasing word
  embeddings}, in: Advances in Neural Information Processing Systems 29, 2016,
  pp. 4349--4357.
\newline\urlprefix\url{https://proceedings.neurips.cc/paper/2016/hash/a486cd07e4ac3d270571622f4f316ec5-Abstract.html}

\bibitem{DBLP:conf/cikm/IosifidisN19}
V.~Iosifidis, E.~Ntoutsi,
  \href{https://doi.org/10.1145/3357384.3357974}{Adafair: Cumulative fairness
  adaptive boosting}, in: W.~Zhu, D.~Tao, X.~Cheng, P.~Cui, E.~A.
  Rundensteiner, D.~Carmel, Q.~He, J.~X. Yu (Eds.), Proceedings of the 28th
  {ACM} International Conference on Information and Knowledge Management,
  {CIKM} 2019, Beijing, China, November 3-7, 2019, {ACM}, 2019, pp. 781--790.
\newblock \href {https://doi.org/10.1145/3357384.3357974}
  {\path{doi:10.1145/3357384.3357974}}.
\newline\urlprefix\url{https://doi.org/10.1145/3357384.3357974}

\bibitem{DBLP:conf/icml/AgarwalBD0W18}
A.~Agarwal, A.~Beygelzimer, M.~Dud{\'{\i}}k, J.~Langford, H.~M. Wallach,
  \href{http://proceedings.mlr.press/v80/agarwal18a.html}{A reductions approach
  to fair classification}, in: Proceedings of the 35th International Conference
  on Machine Learning, Vol.~80 of Proceedings of Machine Learning Research,
  {PMLR}, 2018, pp. 60--69.
\newline\urlprefix\url{http://proceedings.mlr.press/v80/agarwal18a.html}

\bibitem{DBLP:journals/corr/abs-1802-03765}
M.~Olfat, A.~Aswani, \href{http://arxiv.org/abs/1802.03765}{Convex formulations
  for fair principal component analysis}, CoRR abs/1802.03765 (2018).
\newblock \href {http://arxiv.org/abs/1802.03765} {\path{arXiv:1802.03765}}.
\newline\urlprefix\url{http://arxiv.org/abs/1802.03765}

\bibitem{DBLP:conf/innovations/DworkHPRZ12}
C.~Dwork, M.~Hardt, T.~Pitassi, O.~Reingold, R.~S. Zemel,
  \href{https://doi.org/10.1145/2090236.2090255}{Fairness through awareness},
  in: Innovations in Theoretical Computer Science 2012, {ACM}, 2012, pp.
  214--226.
\newblock \href {https://doi.org/10.1145/2090236.2090255}
  {\path{doi:10.1145/2090236.2090255}}.
\newline\urlprefix\url{https://doi.org/10.1145/2090236.2090255}

\bibitem{DBLP:journals/kais/KamiranC11}
F.~Kamiran, T.~Calders, \href{https://doi.org/10.1007/s10115-011-0463-8}{Data
  preprocessing techniques for classification without discrimination}, Knowl.
  Inf. Syst. 33~(1) (2011) 1--33.
\newblock \href {https://doi.org/10.1007/s10115-011-0463-8}
  {\path{doi:10.1007/s10115-011-0463-8}}.
\newline\urlprefix\url{https://doi.org/10.1007/s10115-011-0463-8}

\bibitem{DBLP:conf/sigsoft/ChakrabortyMM21}
J.~Chakraborty, S.~Majumder, T.~Menzies,
  \href{https://doi.org/10.1145/3468264.3468537}{Bias in machine learning
  software: why? how? what to do?}, in: D.~Spinellis, G.~Gousios, M.~Chechik,
  M.~D. Penta (Eds.), {ESEC/FSE} '21: 29th {ACM} Joint European Software
  Engineering Conference and Symposium on the Foundations of Software
  Engineering, Athens, Greece, August 23-28, 2021, {ACM}, 2021, pp. 429--440.
\newline\urlprefix\url{https://doi.org/10.1145/3468264.3468537}

\bibitem{DBLP:conf/cikm/YanKF20}
S.~Yan, H.~Kao, E.~Ferrara, \href{https://doi.org/10.1145/3340531.3411980}{Fair
  class balancing: Enhancing model fairness without observing sensitive
  attributes}, in: M.~d'Aquin, S.~Dietze, C.~Hauff, E.~Curry,
  P.~Cudr{\'{e}}{-}Mauroux (Eds.), {CIKM} '20: The 29th {ACM} International
  Conference on Information and Knowledge Management, Virtual Event, Ireland,
  October 19-23, 2020, {ACM}, 2020, pp. 1715--1724.
\newline\urlprefix\url{https://doi.org/10.1145/3340531.3411980}

\bibitem{DBLP:journals/access/SalazarSAA21}
T.~Salazar, M.~S. Santos, H.~Ara{\'{u}}jo, P.~H. Abreu,
  \href{https://doi.org/10.1109/ACCESS.2021.3084121}{{FAWOS:} fairness-aware
  oversampling algorithm based on distributions of sensitive attributes},
  {IEEE} Access 9 (2021) 81370--81379.
\newblock \href {https://doi.org/10.1109/ACCESS.2021.3084121}
  {\path{doi:10.1109/ACCESS.2021.3084121}}.
\newline\urlprefix\url{https://doi.org/10.1109/ACCESS.2021.3084121}

\bibitem{japkowicz2002class}
N.~Japkowicz, S.~Stephen,
  \href{http://content.iospress.com/articles/intelligent-data-analysis/ida00103}{The
  class imbalance problem: A systematic study}, Intelligent data analysis 6~(5)
  (2002) 429--449.
\newline\urlprefix\url{http://content.iospress.com/articles/intelligent-data-analysis/ida00103}

\bibitem{chawla2002smote}
N.~V. Chawla, K.~W. Bowyer, L.~O. Hall, W.~P. Kegelmeyer,
  \href{https://doi.org/10.1613/jair.953}{Smote: synthetic minority
  over-sampling technique}, Journal of artificial intelligence research 16
  (2002) 321--357.
\newline\urlprefix\url{https://doi.org/10.1613/jair.953}

\bibitem{han2005borderline}
H.~Han, W.-Y. Wang, B.-H. Mao,
  \href{https://doi.org/10.1007/11538059\_91}{Borderline-smote: a new
  over-sampling method in imbalanced data sets learning}, in: International
  conference on intelligent computing, Springer, 2005, pp. 878--887.
\newline\urlprefix\url{https://doi.org/10.1007/11538059\_91}

\bibitem{he2008adasyn}
H.~He, Y.~Bai, E.~A. Garcia, S.~Li,
  \href{https://doi.org/10.1109/IJCNN.2008.4633969}{Adasyn: Adaptive synthetic
  sampling approach for imbalanced learning}, in: 2008 IEEE international joint
  conference on neural networks (IEEE world congress on computational
  intelligence), IEEE, 2008, pp. 1322--1328.
\newline\urlprefix\url{https://doi.org/10.1109/IJCNN.2008.4633969}

\bibitem{douzas2018improving}
G.~Douzas, F.~Bacao, F.~Last,
  \href{https://doi.org/10.1016/j.ins.2018.06.056}{Improving imbalanced
  learning through a heuristic oversampling method based on k-means and smote},
  Information Sciences 465 (2018) 1--20.
\newline\urlprefix\url{https://doi.org/10.1016/j.ins.2018.06.056}

\bibitem{zhou2005training}
Z.-H. Zhou, X.-Y. Liu, \href{https://doi.org/10.1109/TKDE.2006.17}{Training
  cost-sensitive neural networks with methods addressing the class imbalance
  problem}, IEEE Transactions on knowledge and data engineering 18~(1) (2005)
  63--77.
\newline\urlprefix\url{https://doi.org/10.1109/TKDE.2006.17}

\bibitem{sun2007cost}
Y.~Sun, M.~S. Kamel, A.~K. Wong, Y.~Wang,
  \href{https://doi.org/10.1016/j.patcog.2007.04.009}{Cost-sensitive boosting
  for classification of imbalanced data}, Pattern recognition 40~(12) (2007)
  3358--3378.
\newline\urlprefix\url{https://doi.org/10.1016/j.patcog.2007.04.009}

\bibitem{freund1996experiments}
Y.~Freund, R.~E. Schapire, et~al.,
  \href{https://dl.acm.org/doi/10.5555/3091696.3091715}{Experiments with a new
  boosting algorithm}, in: icml, Vol.~96, Citeseer, 1996, pp. 148--156.
\newline\urlprefix\url{https://dl.acm.org/doi/10.5555/3091696.3091715}

\bibitem{freund1997decision}
Y.~Freund, R.~E. Schapire, \href{https://doi.org/10.1006/jcss.1997.1504}{A
  decision-theoretic generalization of on-line learning and an application to
  boosting}, Journal of computer and system sciences 55~(1) (1997) 119--139.
\newline\urlprefix\url{https://doi.org/10.1006/jcss.1997.1504}

\bibitem{DBLP:journals/corr/abs-2105-06345}
E.~Ferrari, D.~Bacciu, \href{https://arxiv.org/abs/2105.06345}{Addressing
  fairness, bias and class imbalance in machine learning: the fbi-loss}, CoRR
  abs/2105.06345 (2021).
\newblock \href {http://arxiv.org/abs/2105.06345} {\path{arXiv:2105.06345}}.
\newline\urlprefix\url{https://arxiv.org/abs/2105.06345}

\bibitem{DBLP:journals/cacm/ChouldechovaR20}
A.~Chouldechova, A.~Roth, \href{https://doi.org/10.1145/3376898}{A snapshot of
  the frontiers of fairness in machine learning}, Commun. {ACM} 63~(5) (2020)
  82--89.
\newblock \href {https://doi.org/10.1145/3376898} {\path{doi:10.1145/3376898}}.
\newline\urlprefix\url{https://doi.org/10.1145/3376898}

\bibitem{du2021fairness}
M.~Du, S.~Mukherjee, G.~Wang, R.~Tang, A.~Awadallah, X.~Hu,
  \href{https://proceedings.neurips.cc/paper/2021/hash/64ff7983a47d331b13a81156e2f4d29d-Abstract.html}{Fairness
  via representation neutralization}, Advances in Neural Information Processing
  Systems 34 (2021) 12091--12103.
\newline\urlprefix\url{https://proceedings.neurips.cc/paper/2021/hash/64ff7983a47d331b13a81156e2f4d29d-Abstract.html}

\bibitem{DOHERTY2012308}
N.~A. Doherty, A.~V. Kartasheva, R.~D. Phillips,
  \href{https://www.sciencedirect.com/science/article/pii/S0304405X12000943}{Information
  effect of entry into credit ratings market: The case of insurers' ratings},
  Journal of Financial Economics 106~(2) (2012) 308--330.
\newblock \href {https://doi.org/https://doi.org/10.1016/j.jfineco.2012.05.012}
  {\path{doi:https://doi.org/10.1016/j.jfineco.2012.05.012}}.
\newline\urlprefix\url{https://www.sciencedirect.com/science/article/pii/S0304405X12000943}

\bibitem{feldman2015computational}
M.~Feldman,
  \href{https://scholarship.tricolib.brynmawr.edu/handle/10066/17628}{Computational
  fairness: Preventing machine-learned discrimination}, Ph.D. thesis (2015).
\newline\urlprefix\url{https://scholarship.tricolib.brynmawr.edu/handle/10066/17628}

\bibitem{DBLP:journals/tkde/HeG09}
H.~He, E.~A. Garcia, \href{https://doi.org/10.1109/TKDE.2008.239}{Learning from
  imbalanced data}, {IEEE} Trans. Knowl. Data Eng. 21~(9) (2009) 1263--1284.
\newblock \href {https://doi.org/10.1109/TKDE.2008.239}
  {\path{doi:10.1109/TKDE.2008.239}}.
\newline\urlprefix\url{https://doi.org/10.1109/TKDE.2008.239}

\bibitem{bunkhumpornpat2009safe}
C.~Bunkhumpornpat, K.~Sinapiromsaran, C.~Lursinsap,
  \href{https://doi.org/10.1007/978-3-642-01307-2\_43}{Safe-level-smote:
  Safe-level-synthetic minority over-sampling technique for handling the class
  imbalanced problem}, in: Pacific-Asia conference on knowledge discovery and
  data mining, Springer, 2009, pp. 475--482.
\newline\urlprefix\url{https://doi.org/10.1007/978-3-642-01307-2\_43}

\bibitem{abdi2015combat}
L.~Abdi, S.~Hashemi, \href{https://doi.org/10.1109/TKDE.2015.2458858}{To combat
  multi-class imbalanced problems by means of over-sampling techniques}, IEEE
  transactions on Knowledge and Data Engineering 28~(1) (2015) 238--251.
\newline\urlprefix\url{https://doi.org/10.1109/TKDE.2015.2458858}

\bibitem{feng2021investigation}
S.~Feng, J.~Keung, X.~Yu, Y.~Xiao, M.~Zhang,
  \href{https://doi.org/10.1016/j.infsof.2021.106662}{Investigation on the
  stability of smote-based oversampling techniques in software defect
  prediction}, Information and Software Technology 139 (2021) 106662.
\newline\urlprefix\url{https://doi.org/10.1016/j.infsof.2021.106662}

\bibitem{DBLP:conf/iclr/ZhangCDL18}
H.~Zhang, M.~Ciss{\'{e}}, Y.~N. Dauphin, D.~Lopez{-}Paz,
  \href{https://openreview.net/forum?id=r1Ddp1-Rb}{mixup: Beyond empirical risk
  minimization}, in: 6th International Conference on Learning Representations,
  {ICLR} 2018, Vancouver, BC, Canada, April 30 - May 3, 2018, Conference Track
  Proceedings, OpenReview.net, 2018.
\newline\urlprefix\url{https://openreview.net/forum?id=r1Ddp1-Rb}

\bibitem{Dua:2019}
D.~Dua, C.~Graff, \href{http://archive.ics.uci.edu/ml}{{UCI} machine learning
  repository} (2017).
\newline\urlprefix\url{http://archive.ics.uci.edu/ml}

\bibitem{janosi1988heart}
A.~Janosi, W.~Steinbrunn, M.~Pfisterer, R.~Detrano,
  \href{https://archive.ics.uci.edu/ml/datasets/Heart+Disease}{Heart disease
  data set} (1988).
\newline\urlprefix\url{https://archive.ics.uci.edu/ml/datasets/Heart+Disease}

\bibitem{propublica2018}
ProPublica,
  \href{https://www.propublica.org/datastore/dataset/compas-recidivism-risk-score-data-and-analysis}{Compas
  recidivism risk score data and analysis} (2018).
\newline\urlprefix\url{https://www.propublica.org/datastore/dataset/compas-recidivism-risk-score-data-and-analysis}

\bibitem{DBLP:journals/jmlr/PedregosaVGMTGBPWDVPCBPD11}
F.~Pedregosa, G.~Varoquaux, A.~Gramfort, V.~Michel, B.~Thirion, O.~Grisel,
  M.~Blondel, P.~Prettenhofer, R.~Weiss, V.~Dubourg, J.~VanderPlas, A.~Passos,
  D.~Cournapeau, M.~Brucher, M.~Perrot, E.~Duchesnay,
  \href{https://dl.acm.org/doi/10.5555/1953048.2078195}{Scikit-learn: Machine
  learning in python}, J. Mach. Learn. Res. 12 (2011) 2825--2830.
\newblock \href {https://doi.org/10.5555/1953048.2078195}
  {\path{doi:10.5555/1953048.2078195}}.
\newline\urlprefix\url{https://dl.acm.org/doi/10.5555/1953048.2078195}

\end{thebibliography}





\end{document}